\documentclass[letterpaper]{article} % DO NOT CHANGE THIS
\usepackage{aaai2026}  % DO NOT CHANGE THIS
\usepackage{times}  % DO NOT CHANGE THIS
\usepackage{helvet}  % DO NOT CHANGE THIS
\usepackage{courier}  % DO NOT CHANGE THIS
\usepackage[hyphens]{url}  % DO NOT CHANGE THIS
\usepackage{graphicx} % DO NOT CHANGE THIS
\usepackage{natbib}  % DO NOT CHANGE THIS AND DO NOT ADD ANY OPTIONS TO IT
\usepackage{caption} % DO NOT CHANGE THIS AND DO NOT ADD ANY OPTIONS TO IT
\usepackage{algorithm}
\usepackage{algorithmic}
\usepackage{pifont}
\newcommand{\cmark}{\ding{51}}  % 打勾
\newcommand{\xmark}{\ding{55}}  % 叉号
\usepackage{booktabs}     % 提供 \toprule, \midrule, \bottomrule
\usepackage{amssymb}      % 提供 \checkmark（可选）

\usepackage{needspace}    % 提供 \needspace，避免分页导致段落断裂
\usepackage{longtable}    % 提供支持跨页的长表格 longtable 环境
\usepackage{booktabs}     % 提供 \toprule, \midrule 等表格美化命令

\usepackage{pifont}       % 提供 \cmark 和 \xmark
\usepackage{graphicx}     % 通用图片支持（如果图中用到）
\usepackage{caption}      % 美化 caption（可选）
\usepackage{lscape}       % 如果你想旋转横向表格（可选）
\usepackage{multirow}
\usepackage{newfloat}
\usepackage{listings}
\DeclareCaptionStyle{ruled}{labelfont=normalfont,labelsep=colon,strut=off} % DO NOT CHANGE THIS
\floatstyle{ruled}
\newfloat{listing}{tb}{lst}{}
\floatname{listing}{Listing}
\title{VideoCogQA: A Controllable Benchmark for Evaluating Cognitive Abilities in Video-Language Models}
\author{
    Chenglin Li \quad Qianglong Chen \quad Zhi Li \quad Feng Tao \quad Yin Zhang\\
    Zhejiang University\\
    Hangzhou, China\\
    {\tt\small \{lichenglin, chenqianglong, lizhi, fengtao, zhangyin\}@zju.edu.cn}
}

\begin{document}

\maketitle

\begin{abstract}
% , which are fundamental to human intelligence
Recent advancements in Large Video-Language Models (LVLMs) have led to promising results in multimodal video understanding. However, it remains unclear whether these models possess the cognitive capabilities required for high-level tasks, particularly those involving symbolic and abstract perception. Existing benchmarks typically rely on real-world, annotated videos, which lack control over video content and inherent difficulty, limiting their diagnostic power. To bridge this gap, we propose \textbf{VideoCogQA}, a scalable and fully controllable benchmark inspired by game-world environments, designed to evaluate the cognitive abilities of LVLMs. By generating synthetic videos via a programmatic engine, VideoCogQA allows fine-grained control over visual elements, temporal dynamics, and task difficulty. This approach enables a focused evaluation of video cognitive abilities, independent of prior knowledge from visual scene semantics. The dataset includes 800 videos and 3,280 question-answer pairs, featuring tasks related to abstract concepts, symbolic elements, and multimodal integration, with varying levels of difficulty. Experimental results show that even state-of-the-art (SOTA) models, such as GPT-4o, achieve an average performance of 48.8\% on tasks involving abstract concepts. Additionally, performance drops by 15\% as task complexity increases, highlighting the challenges LVLMs face in maintaining consistent performance. Through this work, we hope to show the limitations of current LVLMs and offer insights into how they can more effectively emulate human cognitive processes in the future.
\end{abstract}

\begin{figure}
    \centering
    \includegraphics[width=0.9\linewidth]{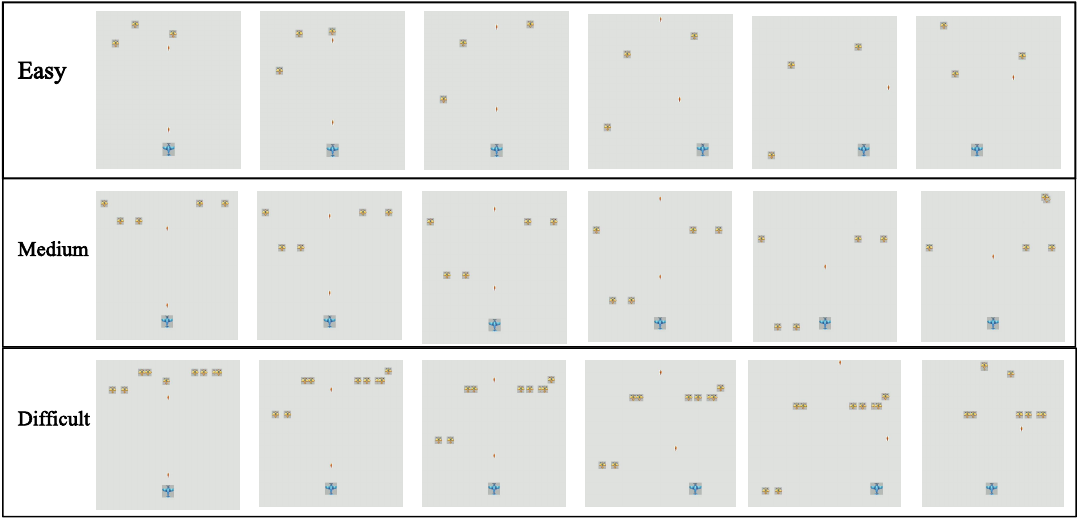}
    \caption{In the \textbf{Sky Battle} scene, symbolic icons are used to represent players, bullets, and enemies, while the action “destroy” is conveyed in an abstract form. Difficulty is controlled by varying the number and speed of enemies. Questions such as “How many enemies are destroyed by player?” are used to test the model’s counting ability in game scenes.}
    \label{SKBCASES}
\end{figure}
\begin{table*}[ht]
\centering
\small
\begin{tabular}{lccccccc}
\toprule
\textbf{Benchmarks} & \textbf{Understanding} & \textbf{Reasoning} & \textbf{Audio} & \textbf{Synthesis} & \textbf{Control} & \textbf{Difficulty Level} \\ 
\midrule
Video-Bench~\cite{ning2023video}      & \cmark & \xmark & \cmark & \xmark & \xmark & \xmark \\ 
MMBench-Video~\cite{fang2024mmbench}  & \cmark & \xmark & \cmark & \xmark & \xmark & \xmark \\ 
AutoEval-Video~\cite{chen2023autoeval} & \cmark & \cmark & \xmark & \xmark & \xmark & \xmark \\ 
MVBench~\cite{li2024mvbench}          & \cmark & \cmark & \xmark & \cmark & \xmark & \xmark \\ 
Video-MME~\cite{fu2024video}          & \cmark & \cmark & \cmark & \xmark & \xmark & \xmark \\
VideoVista~\cite{li2024videovista}    & \cmark & \cmark & \xmark & \cmark & \xmark & \xmark \\ 
VideoCogQA (ours)                     & \cmark & \cmark & \cmark & \cmark & \cmark & \cmark \\ 
\bottomrule
\end{tabular}
\caption{Comparison of video benchmarks across key tasks and characteristics: understanding, reasoning, audio, synthesis (use of generated data), and controllability, along with difficulty level distinction.}
\label{tab:compare_bench}
\end{table*}

\section{Introduction}
\begin{figure}
    \centering
    \includegraphics[width=1.0\linewidth]{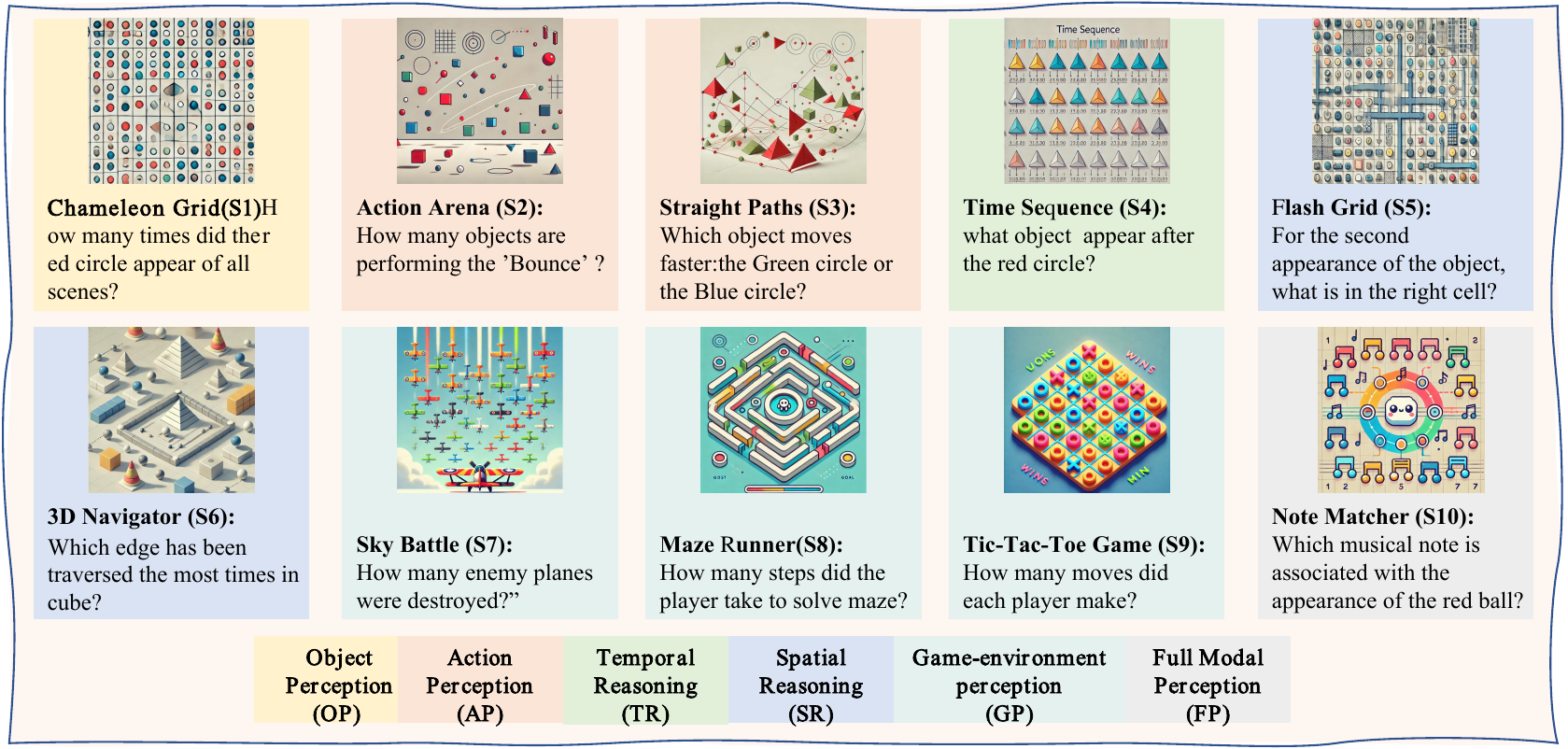}
    % {figure_1.pdf}
    \caption{Overview of VideoCogQA: Task Scenes, Aligned Questions, and Six Core Cognitive Abilities in video scenes.}
    \label{figs:framework}
\end{figure}

The rapid development of artificial intelligence (AI) has driven significant progress in LVLMs, enhancing their ability to process and interpret video data~\cite{li2023videochat,zhang2023video,lin2023video,li2024llava,ye2024mplug,tang2023video}. However, it remains unclear how LVLMs can emulate human-level general intelligence and cognitive abilities, such as symbolic understanding, abstract reasoning, and generalization~\cite{tian2017towards,hagendorff2023human}. While recent benchmarks for large language and vision models have begun incorporating cognition-oriented evaluations~\cite{song2024m3gia, coda2024cogbench, chia2024puzzlevqa}, existing benchmarks for LVLM~\cite{yu2019activitynet,ning2023video,chen2023autoeval,fang2024mmbench,li2024videovista,fu2024video,li2024mvbench} focus mainly on semantic understanding, relying on web-crawled data that lack content control and scalability of video. As a result, symbolic understanding and abstract reasoning is often evaluated implicitly, without directly testing core cognitive abilities. We aim to investigate how LVLMs perceive and interpret video content, generalize from symbolic and abstract elements about object properties such as size, color, and shape, as well as dynamic attributes like motion type and speed, and higher-level spatial and temporal relationships.
% 下次优化方向：可以提供场景语义对 action perception感知 的评测的影响，即提供这个场景但是没有action. 模型也能识别答案
To this end, we propose VideoCogQA, a controllable and scalable benchmark designed to rigorously assess the cognitive capabilities of LVLMs. It utilizes a fully programmatic video synthesis framework, providing fine-grained control over video content, difficulty levels, and task variations. Inspired by classic and popular games such as maze navigation, sky battles, and others, we designed a series of scenes to evaluate key cognitive dimensions in LVLMs. These include Object Perception~\cite{spelke1990principles}, Action Perception~\cite{kelso2018action}, Spatial Reasoning~\cite{malik1983reasoning,stock1998spatial}, Temporal Reasoning~\cite{,mark2020cognitive}, and understanding within Gaming and Full-modal environments~\cite{oei2013enhancing,spence2010video,cohn2016multimodal} as shown in Figure~\ref{figs:framework}. A key advantage of synthetic video-based evaluation is its ability to precisely and scalably assess core abilities across modalities. For example, we can test the model’s ability to perceive actions by observing its interpretation of the motion of symbolic objects (e.g, bouncing, rotating, horizontal movement, etc.), without relying on prior knowledge from contextual cues (a kitchen scene implies the cooking action). Table \ref{tab:compare_bench} provides a comparative analysis of VideoCogQA and existing benchmarks. Through our evaluation of popular LVLMs, we observe that while many models perform well on simple video tasks, their capabilities degrade notably as task complexity increases. For instance, GPT-4o shows a 4\% performance drop when additional objects are introduced in the Action Arena scene, followed by a  10\% decline at the highest difficulty level. Further analysis suggests that the performance drop in temporal tasks stems from the visual encoder’s insufficient ability to grasp high-level abstract and symbolic concepts. These findings underscore the inherent limitations of current models in video-based cognitive tasks and highlight the need for stronger generalization and robustness. Hence, our main contributions are as follows:

\begin{itemize}
    \item We propose a novel video synthesis pipeline using Python that enables the cost-effective generation of video content for capability testing. Integrate GPT-4 designed QA templates and Python-based video generation, with code logs to create batched QA pairs.
    \item To rigorously assess the cognitive abilities of LVLMs, we present \textbf{VideoCogQA}, a scalable and controllable benchmark that uses the automated data synthesis pipeline to evaluate LVLMs in a variety of scene tasks and cognitive dimensions inspired by video games.
    \item Our experiments reveal that even advanced LVLMs struggle with generalization, especially in abstract visual perception, highlighting the need for improved generalization performance in handling high-difficulty tasks.
\end{itemize}

\section{Related Work}
\subsection{Video-LMMs and Benchmark}  
Recent advancements in large multi-modal models~\cite{zhang2023internlm,liu2024visual,liu2024llavanext,Qwen2VL} have greatly enhanced understanding and reasoning capabilities across various domains, especially in image-based tasks~\cite{wu2023q,Fu2023MMEAC,zhang2024benchmark}. As multi-modal research continues to evolve, there is a growing shift from static images to dynamic temporal video~\cite{li2024llms}.  Early investigations in video understanding for LMMs, employing visual encoders, have shown promising results~\cite{li2023videochat,zhang2023video,lin2023video,xu2023retrieval,li2024llms,song2024moviechat,li2024llava,ye2024mplug}.
Meanwhile, active research has focused on constructing benchmarks to assess LVLM capabilities~\cite{ning2023video, chen2023autoeval, fang2024mmbench,li2024videovista, fu2024video, li2024mvbench}. For instance, MVBench~\cite{li2024mvbench} provides a suite of task-specific videos covering various tasks, marking substantial progress in video comprehension. MMBench-Video~\cite{fang2024mmbench} uses extended videos from YouTube and applies free-form questioning to simulate real-world video understanding tasks. However, most existing video-based benchmarks focus on human behavior and contextual understanding, often neglecting abstract cognitive tasks. In MVBench \cite{li2024mvbench}, for instance, when evaluating models' action perception abilities, prior knowledge from the video, such as a playground scene, making it easier to infer the action of running, can lead to shortcut learning. In contrast, our setting uses abstract objects that perform actions such as bouncing or rotating, requiring true motion perception. Moreover, the limited scalability of these benchmarks restricts their broader applicability. To address these limitations, we introduce VideoCogQA, a scalable and controllable dataset to assess a range of cognitive abilities.
\begin{figure*}
    \centering
    \includegraphics[width=1.0\linewidth]{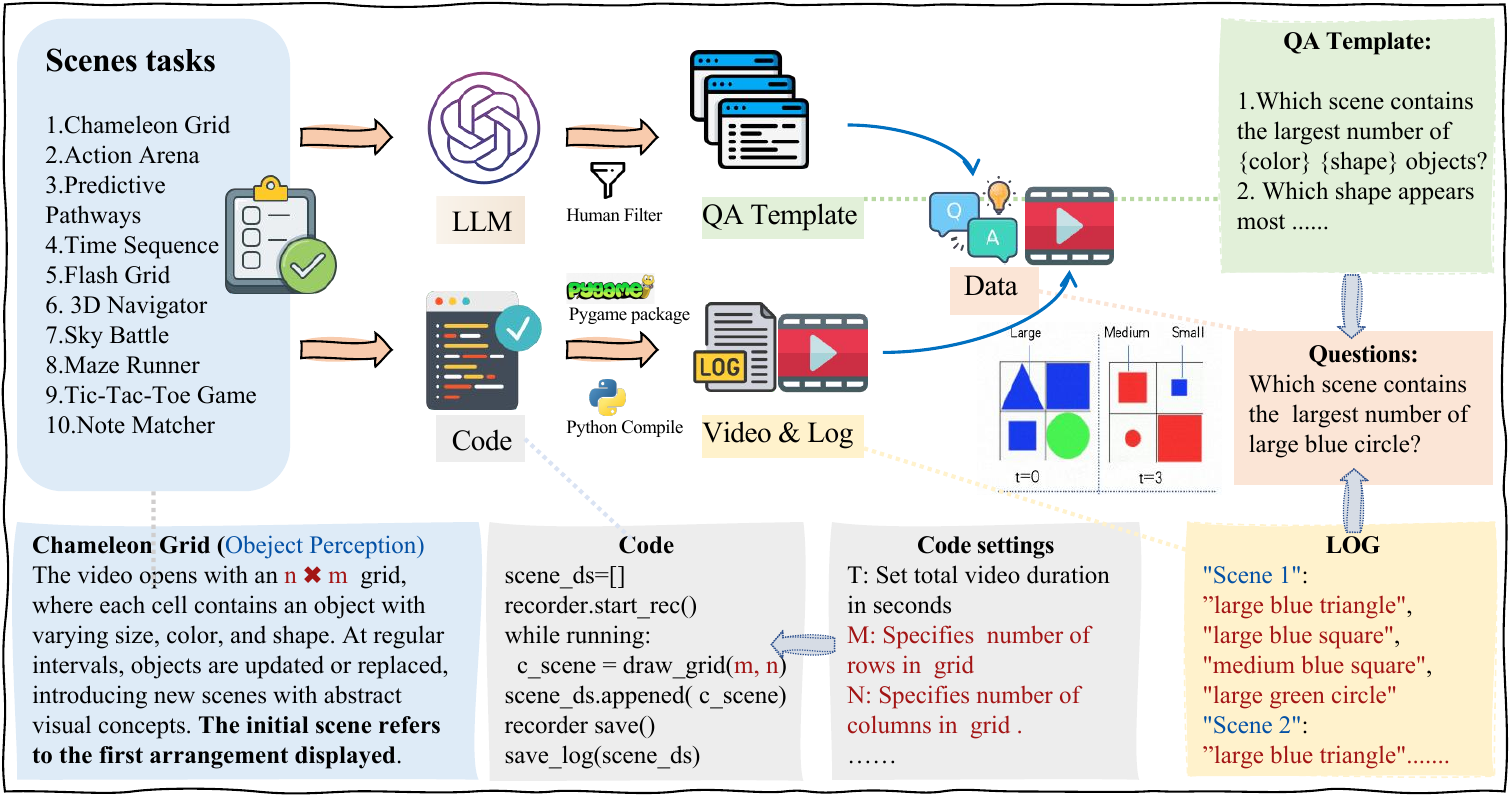}
    \caption{Pipeline for generating videos and corresponding QA templates: Variables \( m \) and \( n \) control the complexity of video scenes. A Python program runs and logs these variables. GPT4 generates scene-related question templates, refined through human filtering. Finally, the variables, QA pairs, and videos are collected.
}
    \label{overview of the pipeline}
\end{figure*}

\subsection{Synthetic Dataset}
Synthetic datasets are cost-effective and avoid the practical challenges of manual annotation~\cite{grauman2022ego4d, chen2023autoeval}, extensive prompt engineering~\cite{li2024mvbench}, and risks of data leakage from pre-trained video corpora~\cite{xu2024benchmarking}. Furthermore, synthetic benchmarks offer a controlled and scalable approach to the evaluation of AI models~\cite{peng2024synthesize, zhao2024needle}. In language model evaluation, synthetic data~\cite{maheshwari2024efficacy} has been used to create structured benchmarks. Similarly, in visual-language model research, synthetic images have been used in controlled experiments to systematically evaluate the visual reasoning~\cite{johnson2017clevr, hudson2019gqa, peng2024synthesize}. Notably, the Abstraction and Reasoning Corpus (ARC)~\cite{chollet2019measure} utilizes programmatically generated images to assess artificial general intelligence. In the video domain, early studies such as Cater~\cite{girdhar2019cater} and Clevrer~\cite{yi2019clevrer} leveraged 3D rendering engines~\cite{blender2018blender} to generate synthetic videos. More recently, synthetic video-based benchmarks have evolved to incorporate multimodal elements. For example, VideoNIAH~\cite{zhao2024needle} integrates textual and visual components into videos to evaluate comprehension. We adopt a Python-based video synthesis framework to construct a scalable and controllable dataset for the evaluation of LVLM. Its synthetic nature enables in-depth analysis of understanding and reasoning abilities beyond existing datasets.

\section{VideoCogQA}

\begin{table*}[h!]
\centering
\begin{tabular}{p{4cm} p{5cm} ccc}
\toprule
\textbf{Scene Name} &\textbf{Parameter Explanation} & \textbf{Easy} & \textbf{Medium} & \textbf{Hard} \\
\midrule
\textbf{Chameleon Grid} & 
\begin{tabular}[c]{@{}l@{}} 
I: Number of cells per row (m) \\ 
J: Number of cells per column (n) \\ 
\end{tabular} & 
\begin{tabular}[c]{@{}c@{}} 
I=2 \\ 
J=2 \\ 
\end{tabular} & 
\begin{tabular}[c]{@{}c@{}} 
I=5 \\ 
J=5 \\ 
\end{tabular} & 
\begin{tabular}[c]{@{}c@{}} 
I=8 \\ 
J=8 \\ 
\end{tabular} \\
\cline{1-5}

\textbf{Action Arena} & 
\begin{tabular}[c]{@{}l@{}} 
N: Number of objects \\ 
A: Number of action types
\end{tabular} & 
\begin{tabular}[c]{@{}c@{}} 
N=3 \\ 
A=3 \\ 
\end{tabular} & 
\begin{tabular}[c]{@{}c@{}} 
N=6 \\ 
A=6 
\end{tabular} & 
\begin{tabular}[c]{@{}c@{}} 
N=9 \\ 
A=8 \\ 
\end{tabular} \\
\cline{1-5}

\textbf{Straight Paths} & 
\begin{tabular}[c]{@{}l@{}} 
N: Number of objects \\ 
A: Number of speed types \\ 
\end{tabular} & 
\begin{tabular}[c]{@{}c@{}} 
N=3 \\ 
A=3 \\ 
\end{tabular} & 
\begin{tabular}[c]{@{}c@{}} 
N=6 \\ 
A=5 \\ 
\end{tabular} & 
\begin{tabular}[c]{@{}c@{}} 
N=9 \\ 
A=8 \\ 
\end{tabular} \\
\cline{1-5}

\textbf{Time Sequence} & 
\begin{tabular}[c]{@{}l@{}} 
T: Time intervals of object changes \\ 
N: Number of  objects \\ 

\end{tabular} & 
\begin{tabular}[c]{@{}c@{}} 
T=5 \\ 
N=3 \\ 
\end{tabular} & 
\begin{tabular}[c]{@{}c@{}} 
T=3 \\ 
N=5 \\ 
\end{tabular} & 
\begin{tabular}[c]{@{}c@{}} 
T=1 \\ 
N=8 \\ 
\end{tabular} \\
\cline{1-5}

\textbf{Flash Grid} & 
\begin{tabular}[c]{@{}l@{}} 
I: Number of cells per row (m) \\ 
J: Number of cells per column (n) \\ 
\end{tabular} & 
\begin{tabular}[c]{@{}c@{}} 
I=2 \\ 
J=2 \\ 
\end{tabular} & 
\begin{tabular}[c]{@{}c@{}} 
I=5 \\ 
J=5 \\ 
\end{tabular} & 
\begin{tabular}[c]{@{}c@{}} 
I=8 \\ 
J=8 \\ 
\end{tabular} \\
\cline{1-5}

\textbf{3D Navigator} & 
\begin{tabular}[c]{@{}l@{}} 
T: Time to traverse each edge \\ 
E: Number of edges \\ 
\end{tabular} & 
\begin{tabular}[c]{@{}c@{}} 
T=2 \\ 
E=5 
\end{tabular} & 
\begin{tabular}[c]{@{}c@{}} 
T=1 \\ 
E=8 
\end{tabular} & 
\begin{tabular}[c]{@{}c@{}} 
T=0.5 \\
E=12 
\end{tabular} \\ 
\cline{1-5}

\textbf{Sky Battle} & 
\begin{tabular}[c]{@{}l@{}} 
N: Number of enemy planes \\ 
A: Number of enemies Speed \\ 
\end{tabular} & 
\begin{tabular}[c]{@{}c@{}} 
N=3 \\ 
A=2 \\ 
\end{tabular} & 
\begin{tabular}[c]{@{}c@{}} 
N=5 \\ 
A=5 \\ 
\end{tabular} & 
\begin{tabular}[c]{@{}c@{}} 
N=10 \\ 
A=8 \\ 
\end{tabular}   \\
\cline{1-5}

\textbf{Maze Runner} & 
\begin{tabular}[c]{@{}l@{}} 
I: Maze length \\ 
J: Maze width \\ 
\end{tabular} & 
\begin{tabular}[c]{@{}c@{}} 
I=3 \\ 
J=3 \\ 
\end{tabular} & 
\begin{tabular}[c]{@{}c@{}} 
I=5 \\ 
J=5 \\  
\end{tabular} & 
\begin{tabular}[c]{@{}c@{}} 
I=8 \\ 
J=8 \\ 
\end{tabular} \\
\cline{1-5}
\textbf{Note Matcher} & 
\begin{tabular}[c]{@{}l@{}} 
T: Time intervals of object changes \\ 
N: Number of notes 
\end{tabular} & 
\begin{tabular}[c]{@{}c@{}} 
T=5 \\ 
N=3 \\ 
\end{tabular} & 
\begin{tabular}[c]{@{}c@{}} 
T=3 \\ 
N=5 \\ 
\end{tabular} & 
\begin{tabular}[c]{@{}c@{}} 
T=1 \\ 
N=7 \\ 
\end{tabular} \\
\bottomrule
\end{tabular}
\caption{Detailed parameters for different scenes across difficulty levels}
\label{tab:complex_scene_parameters}
\end{table*}

\subsection{Dataset Design}
% Task Definition
As language models evolve, recent research~\cite{li2024videovista} has increasingly focused on evaluating their video cognitive capabilities. The common dimensions of evaluation include Object Perception (OP), Action Perception (AP), Temporal Reasoning (TR), and Spatial Reasoning (SR). In VideoCogQA, we expand this scope of video cognition assessment by introducing two key dimensions: Game-environment Perception (GP) and Full-modal Perception (FP). And the synthesized video scenes incorporate symbolic elements and abstract concepts, containing symbolic objects, abstract attributes (color, shape, and size), abstract actions (action type, action speed, and direction), and spatial (2D and 3D scenes) and temporal relationships at varying levels of task difficulty. The following section provides specific descriptions of each dimension.

\begin{itemize} 
% 强调时序
\item \textbf{Object Perception (OP)}: This dimension involves precise recognition of symbolic objects varying in color, shape, and size~\cite{wang2025elysium}. It requires models to sustain high recognition accuracy across diverse visual and abstract attributes.

\item \textbf{Action Perception (AP)}: This capability evaluates the model's proficiency in interpreting the types of actions performed by symbolic objects~\cite{chen2024pca}, accounting for variations in action speed and direction.

\item \textbf{Temporal Reasoning (TR)}: This dimension assesses the model’s capability in understanding and reasoning through abstract sequences of events in videos~\cite{chu2023timebench, fatemi2024test, cai2024temporalbench}, challenging it to track and interpret temporal relationships accurately.

\item \textbf{Spatial Reasoning (SR)}: This dimension evaluates the model's understanding and reasoning regarding spatial relationships within both 2D and 3D contexts~\cite{wu2024mind, tang2024grasp}, addressing abstract elements such as object positioning, orientation, and relative location within video content.

\item \textbf{Game-environment Perception (GP)}: This dimension focuses on the model's comprehension of simulated game environments involving abstract concepts~\cite{wu2023smartplay, topsakal2024benchmarking}. It evaluates the model's ability to interpret game mechanics, predict player actions, and grasp overall game structure, which is critical for LVLMs in analyzing videos embedded in real-life scenarios.

\item \textbf{Full-modal Perception (FP)}: This dimension assesses the model's ability to integrate and process information across multiple modalities—visual, textual, and auditory~\cite{li2024omnibench}. This cross-modal interaction involving various symbolic objects is essential for advanced applications in video analysis.

\end{itemize}

\subsection{Automated Video and QA Generation}

Guided by formal definitions of video cognitive abilities, we developed a synthetic video generation pipeline using Python, inspired by video game environments. This pipeline renders task scenes that incorporate symbolic elements and abstract concepts, with built-in randomness to ensure variability. Videos are produced in batches, while temporal and spatial complexity are precisely controlled by code parameters. Scene logging combined with paired question templates facilitates targeted evaluation of cognitive abilities. Below, we present detailed descriptions of the scenes for VideoCogQA.

\begin{enumerate}
    \item \textbf{Chameleon Grid (OP-S1)}: This scene features \(i \times j\) grids where each cell holds a random symbolic object with unique attributes: size (small, medium, large), color (red, green, blue), and shape (triangle, circle, square). Objects are periodically updated to simulate dynamic visual stimuli inspired by games like \textit{Bejeweled} and \textit{Candy Crush}. Complexity is controlled by adjusting grid dimensions and testing models' object recognition skills in response to changing arrangements.

    \item \textbf{Action Arena (AP-S2)}: This scene includes \(n\) objects performing \(a\) action types, such as horizontal movement, jumping, scaling, and rotation. Complexity is controlled by adjusting the number of objects and the diversity of actions, testing the model’s ability to distinguish action types in a dynamic environment. 

    \item \textbf{Straight Paths (AP-S3)}: This scene involves \(n\) symbolic objects randomly moving in straight lines, bouncing off walls, and altering direction to maintain linear paths with \(a\) speed type. Complexity is controlled by adjusting the number of objects and range of abstract speeds, testing the model’s ability to estimate action speed, interpret action direction, and predict future positions based on motion trajectories.
% \begin{figure*}
%     \centering
%     \includegraphics[width=1.0\linewidth]{latex/pdf/qq.pdf}
%     \caption{Automatically Generated Questions by GPT-4 for the Sky Battle Scene.}
%     \label{qa_template}
% \end{figure*}
\item \textbf{Time Sequence (TR-S4)}: This scene features random symbolic objects appearing and disappearing at set intervals with a simulated clock display, inspired by games like \textit{Simon} and \textit{Guitar Hero}. Complexity is controlled by adjusting the number of objects \(n\) and the set intervals \(t\), testing the model's ability to track timing and sequence changes in an abstract temporal environment.

    \item \textbf{Flash Grid (SR-S5)}: This scene presents a 2D \(i \times j\) matrix where symbolic objects randomly appear in different cells, inspired by games like \textit{Memory Matrix} and \textit{Whac-A-Mole}. Complexity is controlled by adjusting the matrix size, testing the model’s ability to track and recall transient 2D-spatial positions, and interpreting abstract spatial relationships.

    \item \textbf{3D Navigator (SR-S6)}: This scene presents a 3D environment with symbolic objects such as pyramids and cubes, with a small ball randomly moving along their edges, inspired by gameplay reminiscent of \textit{Super Monkey Ball}. Complexity is controlled by adjusting the ball’s abstract speed \(t\) and the intricacy of its path \(e\), testing the model’s ability to track and predict motion within 3D spatial relationships.
    \item \textbf{Sky Battle (GP-S7)}: This scene presents a horizontal player plane at the bottom of the screen, represented by symbolic icons for planes, bullets, and random enemies, inspired by classic arcade gameplay. Complexity is controlled by adjusting the number \(n\) and speed \(a\) of enemy icons, testing the model’s ability to perceive gameplay environments and interpret the symbolic game mechanics.
    \item \textbf{Maze Runner (GP-S8)}: This scene presents a symbolic object navigating a random \(i \times j\) maze toward a designated goal, inspired by classic puzzle gameplay. Complexity is controlled by adjusting the maze design, testing the model’s ability to perceive gameplay environments, and interpreting symbolic game mechanics.
\begin{figure}
    \centering
    \includegraphics[width=1.0\linewidth]{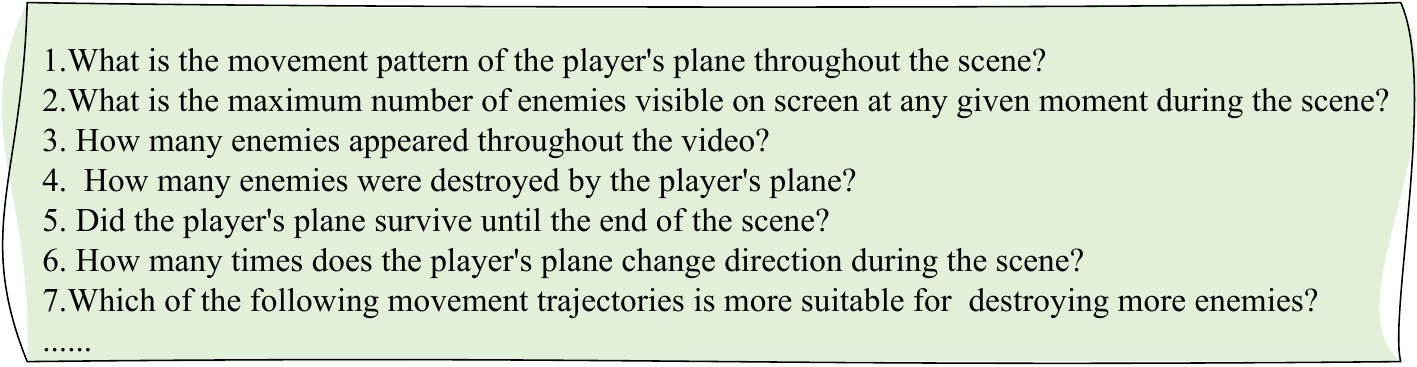}
    \caption{Automatically Generated Questions by GPT-4 in the Sky Battle Scene.}
    \label{qa_template}
\end{figure}

    \item \textbf{Tic-Tac-Toe Game (GP-S9)}: This scene presents a simulated tic-tac-toe game on \(3 \times 3\) grids, testing models' ability to perceive gameplay environments and interpret symbolic game mechanics.

    \item \textbf{Note Matcher (FP-S10)}: This scene presents a single random symbolic object paired with musical notes (1 to 7) inspired by games like \textit{Patapon}. Complexity is controlled by increasing the frequency of object changes \(t\) and note numbers \(n\), testing the model’s ability in audio-visual association and multimodal integration.
\end{enumerate}

We synthesize videos for the specified scenes using Python, allowing fine-grained control over video difficulty by adjusting code parameters, as shown in Figure \ref{overview of the pipeline}. By varying the number of code executions, we can efficiently generate large batches of videos, ensuring scalable evaluation. The GPT-4 prompt used is: ``The above is the code for generating a game video using Pygame. Provide a series of QA templates related to it''. The QA templates in \textbf{Sky Battle} are shown in Figure \ref{qa_template} and code setting is shown in Table ~\ref{tab:complex_scene_parameters}. We employ multiple-choice questions with 3 to 5 shuffled options for automated evaluation. Overall, VideoCogQA comprises 800  generated videos and 3,270 questions. 
% Further details on question templates and code parameter settings are provided in Appendix~\ref{dp}.

\section{Experiments}

\subsection{Setup}
 \label{sec:setup}
We evaluate ten widely used open-source LVLMs fine-tuned on video question-answer pairs, including MiniCPM-V~\cite{yao2024minicpm}, Video-LLaMA2~\cite{damonlpsg2024videollama2}, InternVideo2~\cite{wang2024internvideo2}, Video-LLaVA~\cite{lin2023video}, LLaVA-NEXT-Video-34B~\cite{zhang2024llavanextvideo}, LLaVA-NEXT-Video-7B~\cite{zhang2024llavanextvideo}, and InternLM-XComposer-2.5~\cite{internlmxcomposer2_5}. Additionally, we assess the advanced Qwen2-VL models at different scales, including Qwen2-VL-2B, Qwen2-VL-7B, and Qwen2-VL-72B~\cite{Qwen2VL}, alongside proprietary models, Gemini-1.5-Flash and GPT-4o. Although InternVideo2 can encode audio, we standardize input across all video language models by extracting musical notes per second and converting them into text format. For fairness, all models are evaluated using their default inference settings. The prompts offer descriptions of scene tasks that incorporate abstract visual concepts. 
% Further details on the models can be found in Appendix ~\ref{models}.
% Following the setup~\cite{li2024mvbench}, we use the prefix ‘Best Option:’ in the prompt.
% \begin{figure}
%     \centering
%     \includegraphics[width=0.85\linewidth]{sec/pdf/diff_level.png}
%  \caption{Performance of LVLMs across different levels.}
% \label{level_parameter}

% \end{figure}

% \begin{figure}
%     \centering
%     \includegraphics[width=0.85\linewidth]{latex/pdf/diff_level.png}
%  \caption{Performance of LVLMs across different levels.}
% \label{level_parameter}

% \end{figure}
\begin{figure}
    \centering
    \includegraphics[width=0.85\linewidth]{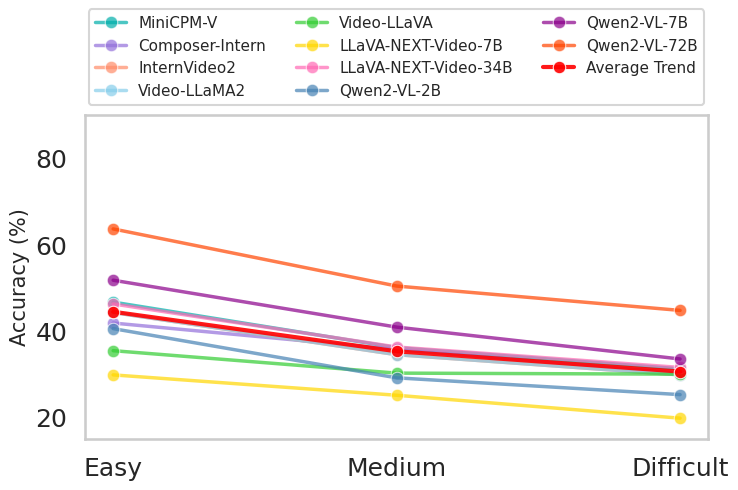}
 \caption{Performance of LVLMs across different levels.}
\label{level_parameter}

\end{figure}

\begin{table*}[ht]
\centering
\setlength\tabcolsep{5.5pt} % 调整列间距
\begin{tabular}{c|c|cc|c|cc|ccc|c}
\toprule
\textbf{Method} & \textbf{OP} & \multicolumn{2}{c|}{\textbf{AP}} & \textbf{TR} & \multicolumn{2}{c|}{\textbf{SR}} & \multicolumn{3}{c|}{\textbf{GP}} & \textbf{FP} \\
\cline{3-4} \cline{6-7} \cline{8-10}
 & \textit{S1} & \textit{S2} & \textit{S3} & \textit{S4} & \textit{S5} & \textit{S6} & \textit{S7} & \textit{S8} & \textit{S9} & \textit{S10} \\
\midrule
\multicolumn{11}{c}{\textbf{Open-Source Models}} \\
\midrule
Random & 33.2 & 34.0 & 37.1 & 30.3 & 32.7 & 23.9 & 25.0 & 28.2 & 37.6 & 33.9 \\
MiniCPM-V & 28.2 & 49.5 & 39.3 & 47.8 & 32.2 & 34.7 & 28.9 & 26.7 & 46.0 & 54.4 \\
Video-LLaMA2 & 31.3 & 50.5 & 33.5 & 48.3 & 36.4 & 18.7 & 26.7 & 27.8 & 52.0 & 52.2\\
InternVideo2 & 31.3 & 50.5 & 33.5 & 48.3 & 36.4 & 18.7 & 26.7 & 27.8 & 52.0 & 52.2\\
Video-LLaVA & 40.4 & 21.0 & 40.8 & 23.2 & 37.5 & 21.3 & 16.7 & 25.5 & 38.0 & 60.0\\
LLaVA-NEXT-Video-7B & 20.4 & 22.5 & 30.7 & 21.0 & 33.8 & 18.7 & 12.2 & 14.4 & 15.3 & 46.7 \\
LLaVA-NEXT-Video-34B & 28.4 & 42.0 & 42.7 & 39.0 & 22.9 & \textbf{58.0} & \textbf{37.8} & 12.2 & 33.3 & 58.9 \\
InternLM-XComposer-2.5  & 36.0 & 38.2 & 45.5 & 44.5 & 43.1 & 20.0 & 8.9 & 25.6 & 35.3 & 61.1  \\
Qwen2-VL-72B & \textbf{51.8} & \textbf{58.2} & \textbf{60.0} & \textbf{56.8} & 60.7 & 42.0 & 32.2 & 37.8 & 62.0 & \textbf{76.7}  \\
\midrule
\multicolumn{11}{c}{\textbf{Closed-Source Models}} \\
\midrule
Gemini-1.5-Flash & 41.3 & 43.3 & 51.5 & 39.8 & 49.1 & 38.7 & 30.0 & 45.6 & 56.0 & 71.1  \\
GPT-4o & 36.4 & 43.7 & 40.8 & 56.5 & \textbf{62.4} & 40.0 & 38.9 & \textbf{40.0} & \textbf{64.0} & 61.1 \\
\bottomrule
\end{tabular}
\caption{Performance of various LVLMs across different scenes.}
\label{tab:average_performance}
\end{table*}

\subsection{Main Results} \label{sec:results}
\begin{figure}
    \centering
    \includegraphics[width=1.0\linewidth]{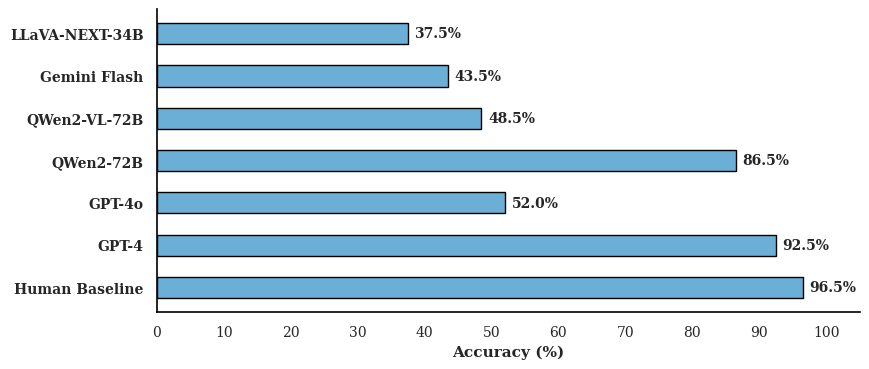}
    \caption{Comparison of human and LVLM performance, with GPT-4 and Qwen2-72B using video descriptions from code logs as substitutes for video.}
    \label{human_compare}
\end{figure}

As shown in Table \ref{tab:average_performance}, most models struggle with OP, SR, and GP tasks, which involve more visual abstract concepts, highlighting their difficulty with advanced video cognition. In contrast, AP and TR primarily test abstract action perception and the temporal reasoning of objects in the scene, with relatively fewer elements. In contrast, their strong performance in FP indicates a solid grasp of integrated audio-visual information when converting musical notes to text. The advanced  Qwen2-VL-72B model stands out among the open-source models tested, consistently achieving the highest accuracy across most tasks, including OP (51.8\%), AP (59.1\%), TR (56.8\%), SR (51.3\%), GP (44.0\%), and FP (76.7\%), leading to an impressive overall average accuracy of 53.7\%. Comparatively, other models like MiniCPM-V show competitive results in AP (44.4\%) and TR (47.8\%), while LLaVA-NEXT-Video-34B excelled in SR (40.4\%) and FP (58.9\%). We also compare model performance to human performance on a 200-sample subset, revealing a significant performance gap shown in Figure~\ref{human_compare}.

 % The InternVideo2 model shows strong results in TR (48.3\%) and FP (52.2\%), whereas Video-LLaVA, despite lower performances in OP (40.4\%) and TR (23.2\%), excels in FP tasks with a high score (60.00\%). While Qwen2-VL-72B excels in video cognition testing with a 53.7\% score, surpassing the random baseline, LVLMs still face challenges, especially in gaming perception tasks with symbolic elements. 
% \begin{figure}
%     \centering
%     \includegraphics[width=1.0\linewidth]{latex/pdf/output.pdf}
%     \caption{Comparison of human and LVLM performance, with GPT-4 and Qwen2-72B using video descriptions from code logs as substitutes for video.}
%     \label{human_compare}
% \end{figure}
% \subsection{Performance Across Different Difficulty Levels}

\subsection{Performance Across Difficulty Levels}
\label{analy1}
As shown in Table~\ref{tab:level_results} and summarized in Figure~\ref{level_parameter}, a fine-grained evaluation reveals that all models exhibit a consistent decline in accuracy as video difficulty increases, underscoring the challenges inherent in complex video cognition tasks. Most models show a roughly 10-point drop from Easy to Medium levels, with an additional 5-point decline at the Difficult level. The evaluation also reveals that, while Video-LLaMA2 and LLaVA-NEXT-Video-34B perform similarly at the Easy level, LLaVA-NEXT-34B begins to outperform Video-LLaMA2 as tasks become more challenging. 
% More results are provided in Appendix \ref{sec:full_results}.
% Under the most challenging conditions, even Qwen2-VL-72B achieves only 23.3\% accuracy in Scene 7 (Sky Battle), approaching the level of random guessing. 
% Similarly, LLaVA-NEXT-Video-34B shows a sharp performance decline at the Difficult level, especially in TR and SR dimensions.  
% These results reveal VLMs' limitations in complex scenes, highlighting the need to improve their generalization in challenging videos.

% \begin{figure}
%     \centering
%     \includegraphics[width=1.0\linewidth]{latex/pdf/parameters.pdf}
%     \caption{Performance of LVLMs under Different Model Parameters.}
%     \label{size_abilities}
% \end{figure}

% %\vspace{-10pt}
% \input{latex/Table/level_results}
\begin{figure}
    \centering
    \includegraphics[width=0.85\linewidth]{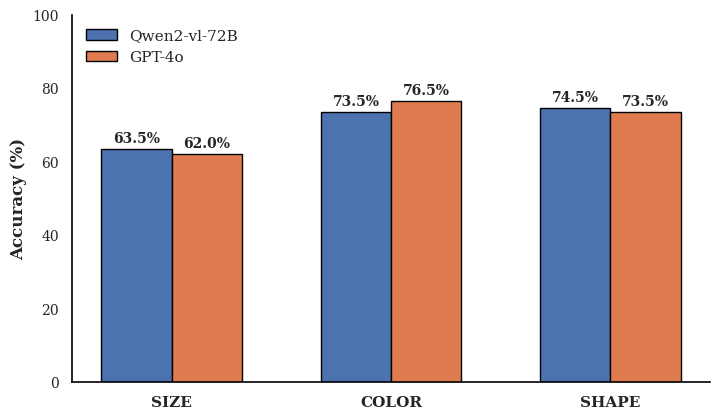}
    \caption{Model Performance on Object Size, Color, and Shape Perception Tasks}
    \label{single_frame_fig}
\end{figure}
\begin{table*}[htbp]
\centering
\setlength\tabcolsep{3pt}
\begin{tabular}{ll|c|cc|c|cc|ccc|c|c}
\toprule
\textbf{Method} & \textbf{Difficulty} & \textbf{OP} & \multicolumn{2}{c|}{\textbf{AP}} & \textbf{TR} & \multicolumn{2}{c|}{\textbf{SR}} & \multicolumn{3}{c|}{\textbf{GP}} & \textbf{FP}  & \textbf{Avg.}\\
\cline{3-13}
& & \textit{S1} & \textit{S2} & \textit{S3} & \textit{S4} & \textit{S5} & \textit{S6} & \textit{S7} & \textit{S8} & \textit{S9} & \textit{S10} \\
\midrule
\multirow{3}{*}{MiniCPM-V}
& Easy     & 34.7 & 54.5 & 50.5 & 53.5 & 47.3 & 36.0 & 43.3 & 30.0 & 46.0 & 70.0 & 46.3 \\
& Medium   & 26.0 & 48.5 & 36.5 & 50.5 & 30.0 & 40.0 & 23.3 & 26.7 & ---   & 43.3 & 35.8 \\
& Difficult& 24.0 & 45.5 & 31.0 & 39.5 & 19.3 & 28.0 & 20.0 & 23.3 & ---   & 50.0 & 31.0\\
\midrule

\multirow{3}{*}{Video-LLaMA2}
& Easy     & 41.3 & 61.0 & 46.5 & 50.0 & 49.3 & 20.0 & 26.7 & 33.3 & 52.0 & 70.0 & 44.8 \\
& Medium   & 29.3 & 52.0 & 26.5 & 53.0 & 31.3 & 18.0 & 23.3 & 31.0 & ---   & 56.7 & 34.2 \\
& Difficult& 23.3 & 38.5 & 27.5 & 42.0 & 28.7 & 18.0 & 30.0 & 30.0 & ---   & 30.0 & 29.6 \\
\midrule

\multirow{3}{*}{LLaVA-NEXT-Video-34B}
& Easy     & 29.3 & 48.5 & 53.5 & 44.0 & 42.0 & 62.0 & 56.7 & 21.0 & 33.3 & 70.0 & 44.7 \\
& Medium   & 26.0 & 40.0 & 42.5 & 45.0 & 18.7 & 54.0 & 36.7 & 16.7  & ---   & 56.7 & 36.9\\
& Difficult& 30.0 & 37.5 & 32.0 & 28.0 & 8.0  & 58.0 & 20.0 & 20.0 & ---   & 50.0 & 31.4\\
\midrule

\multirow{3}{*}{Qwen2-VL-72B}
& Easy     & 61.3 & 65.0 & 67.0 & 64.5 & 76.0 & 52.0 & 40.0 & 63.3 & 62.0 & 83.3  & 63.3\\
& Medium   & 48.7 & 59.5 & 62.5 & 56.0 & 56.0 & 34.0 & 33.3 & 23.3 & ---   & 80.0 & 50.1\\
& Difficult& 45.3 & 50.0 & 50.5 & 50.0 & 50.0 & 40.0 & 23.3 & 26.7 & ---   & 66.7 & 44.4 \\

\bottomrule
\end{tabular}
\caption{Performance of various models across different scenes and difficulty levels}
\label{tab:level_results}
\end{table*}
\subsection{Model Performance Analysis}
We hypothesize that the poor performance of LVLMs on temporal tasks stems from limitations in their visual encoders’ ability to perceive high-level abstract and symbolic concepts. This hypothesis is supported by two lines of analysis. One supporting observation is that replacing videos with full temporal descriptions extracted from code logs—where symbolic and abstract elements are explicitly presented in textual form—leads to substantial performance improvements. This effect is especially pronounced for large models such as Qwen2-72B and GPT-4, as shown in Figure~\ref{human_compare}. The performance gains suggest that models struggle not with reasoning over temporal content itself, but rather with extracting relevant abstract information from raw visual input. 

To further examine the visual encoder’s symbolic perception capability, we leverage the existing Chameleon Grid (S1) task, which evaluates object-level temporal understanding. This task is specifically designed to assess whether models can track and interpret symbolic object properties—such as color, shape, and size—across time and spatial positions. Most models exhibit poor performance on this task. To better understand the underlying limitations, we conduct a descriptive single-frame test using sampled frames from the task videos, which are cognitively simple for humans to interpret and can be easily understood without errors. Specifically, we randomly select 100 frames each from S1 and prompt the models to describe the symbolic elements in the frame based on their grid positions (organized by rows and columns). We then compute the accuracy of the model's responses for each grid location.  As shown in Figure~\ref{single_frame_fig}, both models frequently misidentify object size, revealing limitations in fine-grained perceptual capability. This suggests that widely used visual encoders such as CLIP \cite{radford2021learning} may lack sufficient pretraining on fine-grained symbolic elements, particularly those related to object size.

% Detailed results are provided in Appendix~\ref{single_frame}.

\subsection{Impact of Model Size on Performance}
\label{analy2}
\begin{figure}
    \centering
    \includegraphics[width=1.0\linewidth]{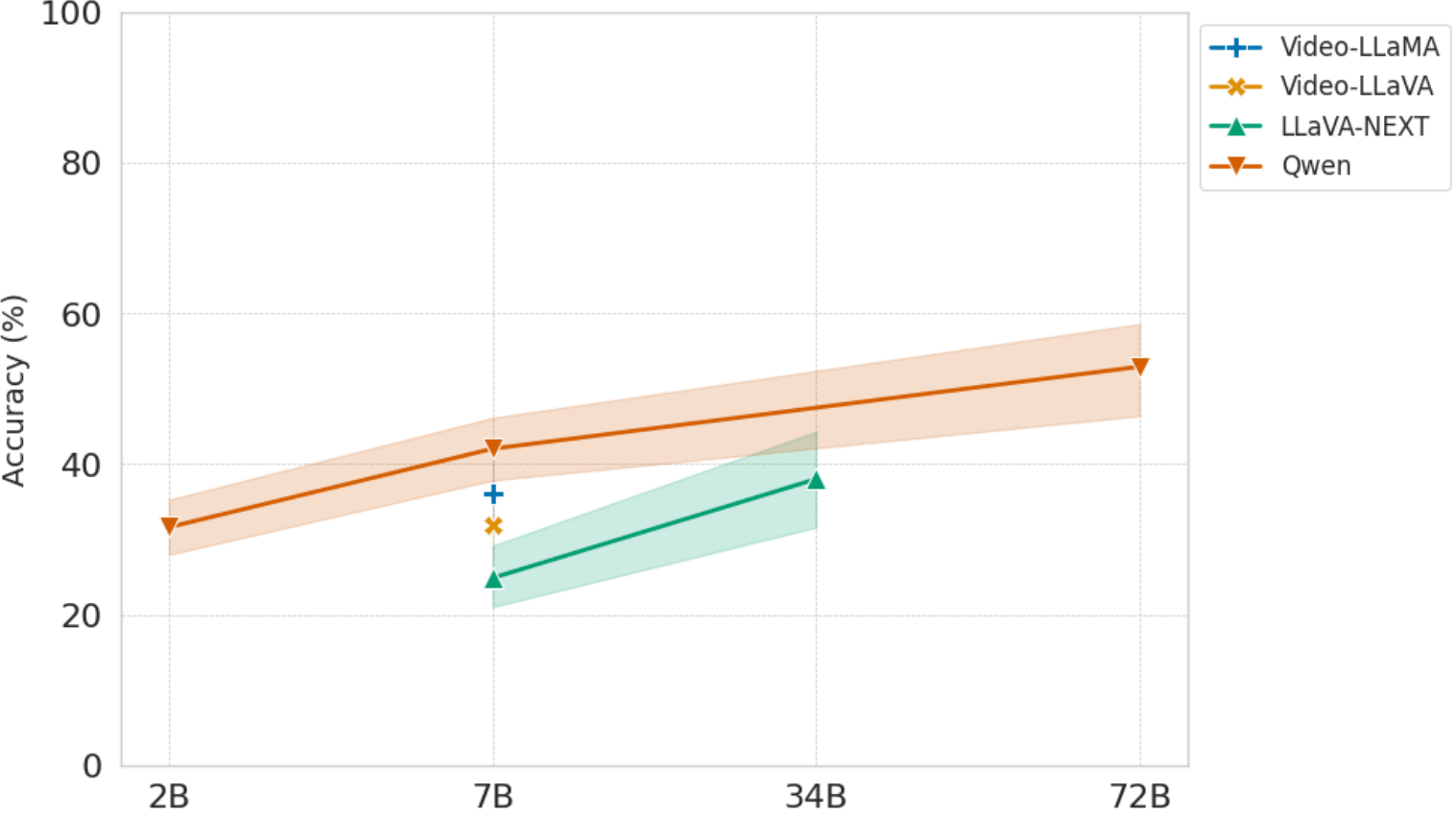}
    \caption{Performance of LVLMs under Different Model Parameters.}
    \label{size_abilities}
\end{figure}

The scale of parameters plays a crucial role in determining the performance of language models~\cite{brown2020language, Qwen2VL}. Figure~\ref{size_abilities} demonstrates a strong positive correlation between model size and performance. For the Qwen model, as the model size scales from 2B to 7B and further to 72B, average performance scores rise significantly, from 31.9 to 42.5 and then to 53.7. 

\begin{figure}
    \centering
    \includegraphics[width=0.9\linewidth]{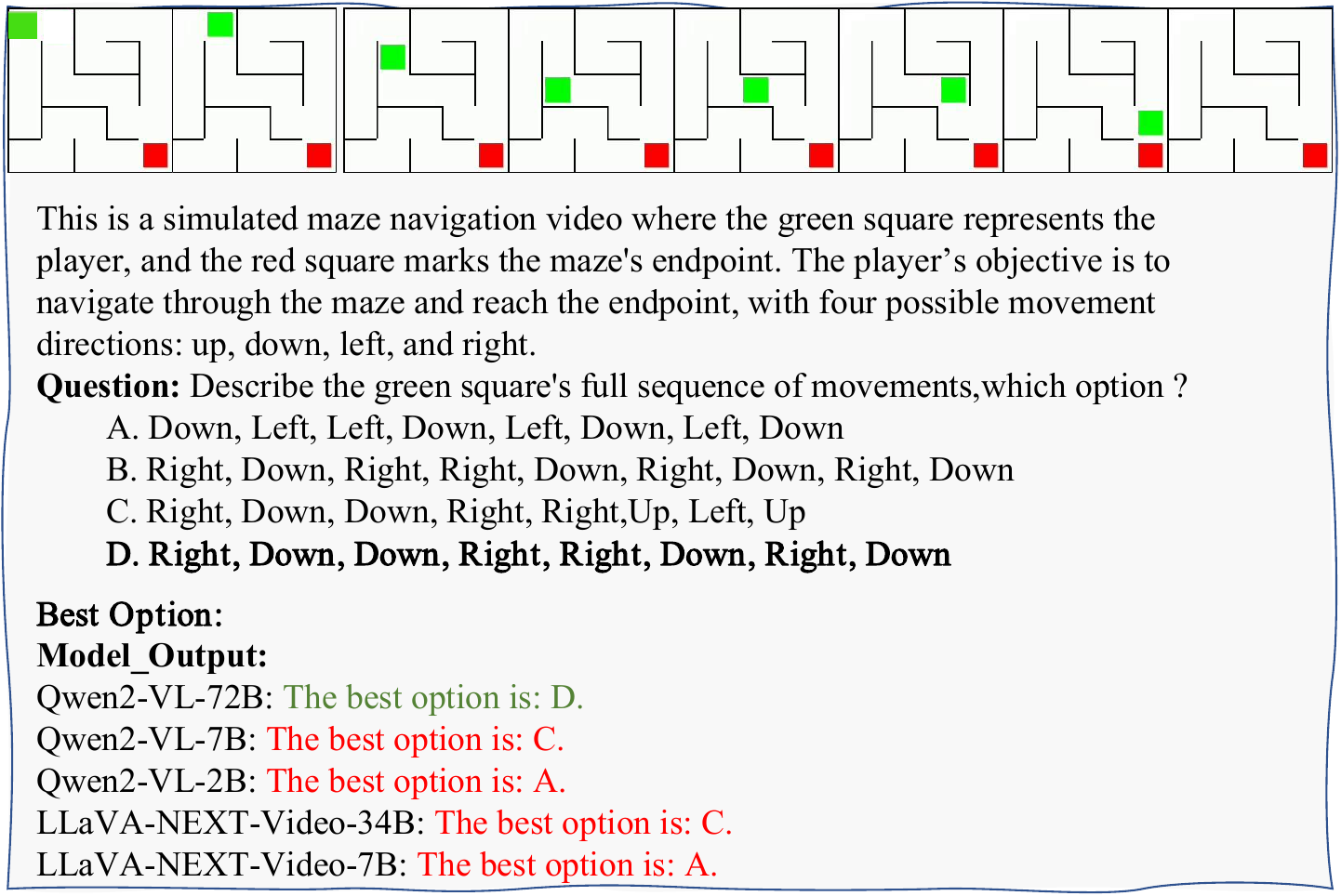}
    \caption{The case in the Maze Navigation scene.}
    \label{fig:maze_case}
    
\end{figure}
\subsection{Case Study}
\label{analy3}
In Figure~\ref{fig:maze_case}, we present a simple case from the Maze Navigation scene. We present a video description task involving the movement trajectory of the green block in the video, which is easy for humans. To answer correctly, the LVLM must accurately perceive the player's spatial trajectory and retain the path structure. Among all models, only Qwen2-VL-72B correctly selected option D, demonstrating a strong understanding of the game environment and successfully identifying the optimal path. In contrast, Qwen2-VL-7B and LLaVA-NEXT-Video-34B chose option C, suggesting a partial understanding of the spatial reasoning task. Meanwhile, Qwen2-VL-2B and LLaVA-NEXT-Video-7B selected option A, indicating an incorrect initial interpretation and reflecting considerably weaker cognitive capabilities.
% in this abstract video setting. 

% Notably, the performance of Qwen2-VL-2B is comparable to that of the larger LLaVA-NEXT-Video-7B, highlighting the efficiency of the Qwen series in handling complex video tasks despite a smaller model size.

% \input{latex/Figures/case11}
\section{Conclusion}

% 打磨结论这一部分内容，强调多样和可可扩展, 泛化性能
In this work, we introduce VideoCogQA, a controllable and scalable evaluation dataset designed to assess the cognitive abilities of LVLMs across diverse video tasks. VideoCogQA allows for precise content alignment and adjustable difficulty tailored to specific cognitive evaluations. Our experiments reveal that even SOTA models, such as GPT-4o and Qwen2-VL-72B, face significant challenges with symbolic elements, with performance dropping sharply as video difficulty increases. These results underscore the need to improve the generalization of cognitive capabilities in LVLMs.

\bibliography{aaai2026}

\end{document}